\documentclass{article}
\usepackage{graphicx} 
\usepackage[utf8]{inputenc}
\usepackage{cite}
\usepackage{amsmath}
\usepackage{amssymb}
\usepackage{multirow}
\usepackage{booktabs, longtable, array}
\usepackage{authblk}
\usepackage{mathrsfs}

\title{Enhancing operational wind downscaling capabilities  over Canada: Application of a Conditional Wasserstein GAN methodology}

\author[1]{Jorge Guevara}
\author[1]{Victor Nascimento}
\author[1]{Johannes Schmude}
\author[1]{Daniel Salles Civitarese}
\author[2]{Simon Corbeil-Létourneau}
\author[2]{Madalina Surcel}
\author[2]{Dominique Brunet}

\affil[1]{IBM Research}
\affil[2]{Environment and Climate Change Canada}

\date{}

\begin{document}

\maketitle

\begin{abstract}
Downscaling based on Artificial Intelligence (AI) has been demonstrated to be a powerful tool for improving the spatial resolution of weather forecasts. 
This study advances wind downscaling by extending the DownGAN framework introduced by Annau et al.\,to operational Canadian Numerical Weather Predictions (NWP), namely from the Global Deterministic Prediction System (GDPS) to the domain of the High-Resolution Deterministic Prediction System (HRDPS). 
We enhance the AI model by incorporating high-resolution static covariates, such as topography on the HRDPS grid, into a Conditional Wasserstein Generative Adversarial Network with Gradient Penalty, implemented using a UNET-based generator. 
Following the DownGAN framework, our methodology integrates low-resolution GDPS forecasts (15 km, 10-day horizon) and high-resolution HRDPS forecasts (2.5 km, 48-hour horizon) with Frequency Separation techniques adapted from computer vision.
Through robust training and inference over the whole Canadian region, we demonstrate the operational scalability of our approach, while achieving significant improvements in wind downscaling accuracy. 
Statistical validation highlights reduction in root mean square error (RMSE) and log spectral distance (LSD) metrics compared to the original DownGAN. High-resolution conditioning covariates and Frequency Separation strategies prove instrumental in enhancing model performance. 
This work underscores the potential for extending high-resolution wind forecasts beyond the 48-hour horizon, bridging the gap to the 10-day low resolution global forecast window.
\end{abstract}

\section{Introduction}

Accurate high-resolution wind forecasts are crucial for several application sectors such as aviation \cite{Wynnyk2012Wind}, wind energy \cite{Chen2024Novel}, power infrastructure \cite{HuangWang2024Adaptive}, public safety \cite{LangEtAl2022Predicting}, wildfire management \cite{Beer1991Interaction}, avalanche warnings \cite{LehningFierz2008Assessment},  winter sports \cite{WangEtAl2024Developing}, air quality \cite{Yassin2013Numerical} and water quality management \cite{BrunetEtAl2023Wind}. Typically, high-resolution wind forecasts are obtained from numerical weather predictions (NWPs), starting with a global forecast model followed by a limited-area forecast model which takes the lower-resolution model as boundary conditions in a procedure called dynamical downscaling. In Canada, high-resolution numerical prediction models have a limited forecast horizon of 48 hours due to computational constraints. An applied perspective of the work presented in this paper is to extend high-resolution   wind forecast services from 48 hours to the same 10 days forecast horizon as the Canadian global forecast model.

In meteorology, downscaling is the operation of passing from a coarse (or low-resolution) field to a fine (or high-resolution) field. Besides dynamical downscaling, statistical downscaling is another well established approach for correcting wind at station locations. Several variations of Model Output Statistics (MOS) \cite{GlahnLowry1972Use, BedardEtAl2013Development} have been proposed in the literature, with the Updateable Model Output Statistics (UMOS) \cite{WilsonVallee2002Canadian} approach the one implemented operationally at Environment and Climate Change Canada (ECCC). All the MOS techniques are essentially based on the concept of Multiple Linear Regression (MLR), where several predictors from the lower-resolution NWP output are combined linearly to minimize the mean square error against an observed weather field, with one specific model used at each station location. MOS techniques can be expanded to gridded fields as long as a high-resolution gridded target is available, or specific corrections applied at each station can be interpolated to the grid \cite{GlahnEtAl2009Gridding}, but they cannot capture complex non-linear spatial interactions between variables.

More recently, several contributions have demonstrated the potential of Artificial Intelligence (AI) based downscaling \cite{AdytiaEtAl2024Deep, AnnauEtAl2023Algorithmic, Bano-MedinaEtAl2020Configuration, DujardinLehning2022WindTopo, DupuyEtAl2023Downscaling, HarrisEtAl2022Generative, HohleinEtAl2020Comparative, HuEtAl2023Downscaling, LeToumelinEtAl2024Twofold, LinEtAl2023Deep, MirallesEtAl2022Downscaling, SekiyamaEtAl2023Surrogate}, applying successfully techniques originally developed for image processing and computer vision to weather data. 

In the field of image processing, the image super-resolution (or zooming) problem consists of taking a coarser image as input and producing a finer image at higher resolution.\footnote{Confusingly, what is called upsampling (downsampling) in the field of image processing is called downscaling (upscaling) in meteorology.} Earlier image super-resolution methods were achieving an enhancement factor of 2 to 4 times, but with the advent of deep learning, an enhancement factor of 8 and more is now possible. Generative AI in particular has proven to be a powerful tool not only to artificially produce realistic-looking images and videos from a random seed, but also to condition the generative AI on coarser or partially missing data (called inpainting).

AI-based techniques for downscaling weather and climate fields can be classified in an increasing order of complexity and recency. At the end of spectrum, ensemble decision trees  \cite{DujardinLehning2022WindTopo} techniques such random forests and extreme gradient boosting are the most similar to MLR and the easiest to implement. Other classical machine learning techniques such as support vector machine and shallow artificial neural networks with a single hidden layer can also fit in that category. Methods based on Convolutional Neural Networks (CNNs) \cite{DupuyEtAl2023Downscaling, SekiyamaEtAl2023Surrogate, GirouxEtAl2024InterpolationFree}, have proved to be both popular and powerful. Generative Adversarial Networks (GANs) are the first type of generative AI allowing to sample an ensemble of possible realizations \cite{AnnauEtAl2023Algorithmic, MirallesEtAl2022Downscaling, DaustMonahan2024Capturing}. More recently, latent diffusion techniques have emerged as an interesting alternative to GAN \cite{TomasiEtAl2024Can}. Finally, methods based on vision transformers and in particular those that consist of fine-tuning a weather foundation model are emerging.

In the literature, these methods have been applied to a variety of weather fields such as temperature, wind speed or wind vectors, and precipitation. Many of the published works focus on one or several small areas as a proof-of-concept, and few are used in an operational setting. It is important to note that similar AI techniques can be used either for downscaling weather forecasts, climate projections or reanalysis data, with the main difference in the choice and resolution of source and target data. 
Moreover, with a general downscaling model, there is no conceptual difference between downscaling to a finer grid or downscaling to a point location that could be either an existing weather station or any location of interest.

The key contributions of this work are as follows: 1) We extend the DownGAN framework introduced by \cite{AnnauEtAl2023Algorithmic} by 
applying its methods to operational Numerical Weather Prediction (NWP) datasets, specifically GDPS and HRDPS 
across the entire Canadian domain, expanding beyond the regional focus of the original study.
2) We enhance the DownGAN model, a Wasserstein GAN with Gradient Penalty (WGAN-GP) framework for wind downscaling, 
by introducing high-resolution static covariates, such as topography from the HRDPS domain, 
as conditioning inputs within the generator architecture, 
implemented using a UNET. 
We refer this modified framework as a Conditional WGAN-GP.
3) We propose a robust training and inference schema that allows the model to effectively 
learn wind patterns and generate predictions over the entire Canadian region. 
This development ensures the scalability and operational relevance of the methodology.
4) We perform statistical validation for large-scale wind downscaling across the Canadian domain, 
identifying limitations such as the ineffectiveness of Annau's partial frequency separation technique in this context.
However, Frequency Separation methods adapted from computer vision and used by DownGAN show strong performance. 
The validation highlights the importance of high-resolution conditioning covariates and the utility of Frequency Separation for downscaling.

The paper is divided as follows. Section 2 introduces the datasets and the pre-processing steps in preparation for training the AI model. Section 3 presents the methods for training and inference of the AI model. Experimental results follow in Section 4, and Discussion and Conclusions are presented in Section 5.

\section{Dataset}
\label{sec:dataset}
\subsection{Numerical Weather Predictions}
Recall that the goal of this study is to extend the forecast horizon of the High-Resolution Deterministic Prediction System (HRDPS) by downscaling the Global Deterministic Prediction System (GDPS).
The Global Deterministic Prediction System (GDPS) \cite{GassetEtAl2021Global} is ECCC's operational long-range forecast model. It represents the global configuration of the Global Environmental Multiscale (GEM) model \cite{CoteEtAl1998Operational} and it is run twice a day (at 00 UTC and 12 UTC), producing 10-day forecasts at a 15-km nominal resolution. 

The High-Resolution Deterministic Prediction System (HRDPS) \cite{MilewskiEtAl2021High} is a limited-area model (LAM) configuration of GEM for Canada and Northern United States. The prediction is run operationally at ECCC every 6 hours (00 UTC, 06 UTC, 12 UTC and 18 UTC runs) for a 48 hours forecast at 2.5-km nominal resolution. The HRDPS is the main source of NWP guidance for the Meteorological Service of Canada for day 1 and day 2 forecasts.

\subsection{Predictors and Predictands}
High-resolution zonal (East-West) and meridional (North-South) surface (10 m) wind components from HRDPS are used as the target variables. By predicting both components of the wind vector, we can derive both wind speed and direction at the resolution of HRDPS. Low-resolution zonal and meridional surface wind from GDPS are taken as predictors along with surface temperature, as well as wind, temperature and vertical motion at 546 hPa. While we desired to stay close to the choices of predictors of \cite{AnnauEtAl2023Algorithmic}, we did not include the Convective Available Energy Potential (CAPE) as it was not readily available in GDPS with an hourly resolution. We instead chose surface and mid-troposphere variables as they are expected to be important drivers of surface winds. However, we did not perform any variable selection experiment.
In contrast with \cite{AnnauEtAl2023Algorithmic}, the following three high-resolution geophysical fields are added as predictors: model orography, land-water mask and surface roughness length. These static high-resolution covariates can be used as predictors since these variables are constant at the time-scale of weather prediction. The model orography field is particularly important in mountainous regions where wind channeling is predominant. The land-water mask not only provides an indication of changing surface roughness between the smoother water and other land, but can also be a driver of local winds such as sea breezes and lake breezes. Finally, the surface roughness variable modulates the speed of the wind, and changes in this covariate can drive turbulence in the planetary boundary layer. 
Table\,\ref{table:vars_description} describes the meteorological variables used for training the AI downscaling model. 


\begin{footnotesize} 
\begin{longtable}{@{}p{1cm} p{10.5cm}@{}}
\label{table:vars_description} \\
\caption{Description of predictors and predictands for the downscaling experiments. The HRDPS static covariates are extra high-resolution predictors used in the Conditional WGAN-GP, whereas the DownGAN baseline we implemented only uses GDPS predictors.} \\ 
\toprule
\textbf{Description} \\ 
\midrule
\multicolumn{2}{c}{\textbf{GDPS Predictors}} \\ 
\midrule
$\mbox{U}_{\mbox{surf}}$ & True geographical West-East (zonal) component of the horizontal wind at the surface (10 m) [m/s]. \\
$\mbox{V}_{\mbox{surf}}$ & True geographical South-North (meridional) component of the horizontal wind at the surface (10 m)  [m/s]. \\
$\mbox{T}_{\mbox{surf}}$ & Air temperature at the surface (1.5 m) [°C]. \\
$\mbox{T}_{546}$ & Air temperature vertically interpolated at 546 hPa  [°C]. \\
$\mbox{U}_{546}$ & True geographical West-East (zonal) component of the horizontal wind vertically interpolated at 546 hPa [m/s]. \\
$\mbox{V}_{546}$ & True geographical South-North (meridional) component of the horizontal wind vertically interpolated at 546 hPa  [m/s]. \\
$\mbox{W}_{546}$ & Vertical motion vertically interpolated at 546 hPa  [Pa/s]. \\ 
\midrule
\multicolumn{2}{c}{\textbf{HRDPS Predictands}} \\ 
\midrule
u10 & True geographical West-East (zonal) component of the horizontal wind at the surface (10 m) [m/s]. \\
v10 & True geographical South-North (meridional) component of the horizontal wind at the surface (10 m) [m/s]. \\ 
\midrule
\multicolumn{2}{c}{\textbf{HRDPS Static Covariates}} \\ 
\midrule
me & Model orography [m]. \\
mg & Water/land mask [fraction]. \\
z0 & Roughness length [m]. \\ 
\bottomrule
\end{longtable}
\end{footnotesize}

\subsection{Data pre-processing}
\label{sec:preprocessing}
\begin{figure}[ht]
    \centering
    \includegraphics[width=1\textwidth]{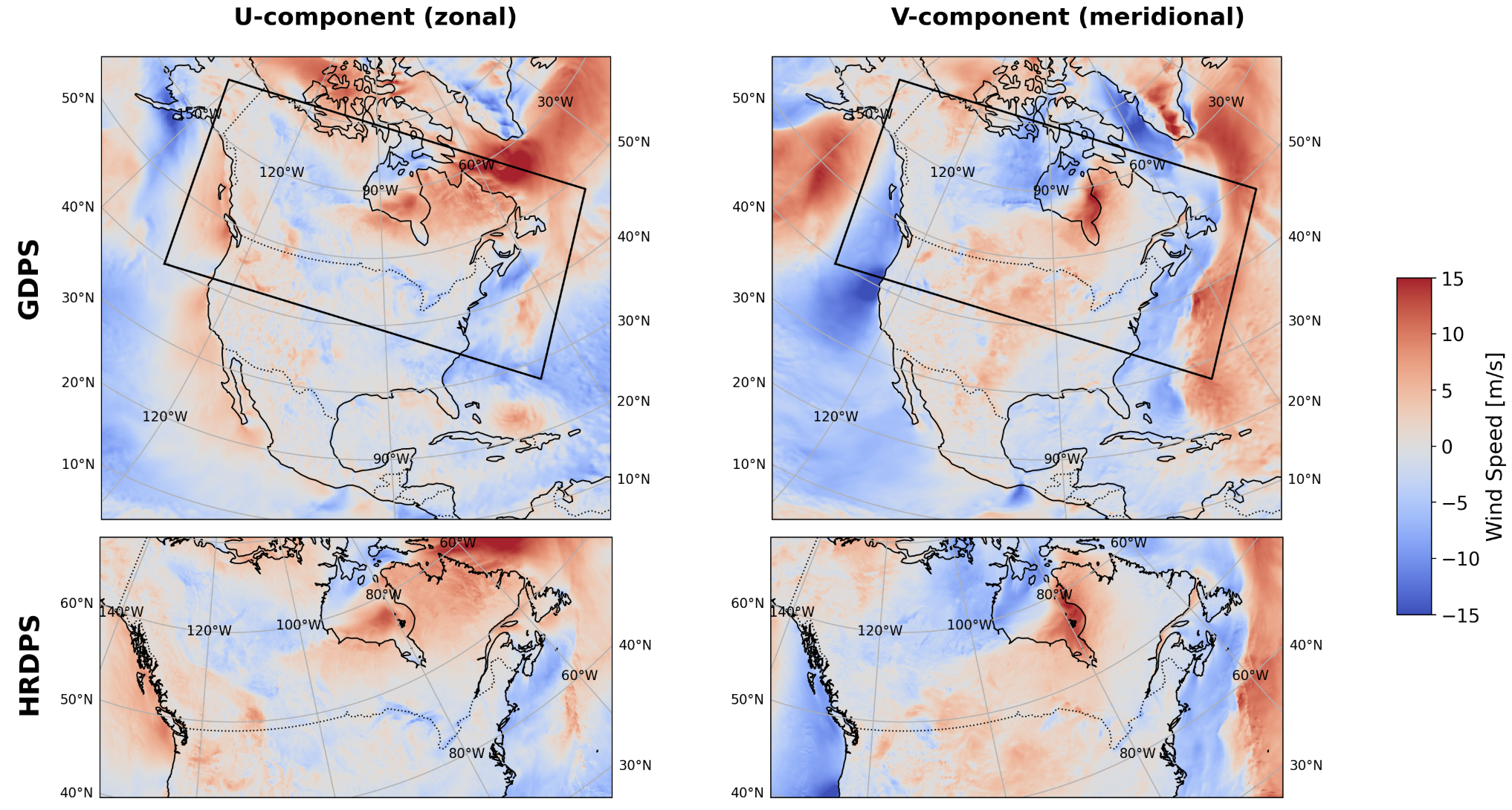}
    \caption{Example of zonal and meridional winds on the GDPS and HRDPS grids. Note that the GDPS grid was cropped over North America. The domain of HRDPS grid is indicated by a black rectangle.}
    \label{fig:gdps_hrdps}
\end{figure}

The map projection of the GDPS outputs is a Yin-Yang grid with a different rotated latitude-longitude map projection than the one of the HRDPS grid. To align the grids, we project\footnote{The regridding process was done using the xESMF Python Library \cite{xesmf}.} the GDPS grid to the HRDPS grid using nearest neighbor interpolation. The interpolated data is then reduced by a factor of 8 (20-km nominal resolution) as the input of the AI downscaling method. Fig.\,\ref{fig:gdps_hrdps} illustrates the respective domains of GDPS (cropped over North-America) and of HRDPS.

\section{Methods}
In this study, we adapt the deep learning framework for downscaling wind components proposed by \cite{AnnauEtAl2023Algorithmic}. This framework is based on Wasserstein Generative Adversarial Networks with Gradient Penalty (WGAN-GP) \cite{GulrajaniEtAl2017Improved} and incorporates elements from Generative Adversarial Networks for image super-resolution (SRGAN) \cite{LedigEtAl2017Photorealistic} as a baseline for the downscaling task.  In the following sections, we elaborate on the application of WGANs for the downscaling problem, describe the model architecture, and introduce our approach for integrating high-resolution static covariates into the training process using UNETs \cite{RonnebergerEtAl2015Unet}.

\subsection{Conditional WGAN-GP for Downscaling}
We consider a training set consisting of dynamic low-resolution predictors, such as GDPS fields, and high-resolution predictands, such as HRDPS fields, with matched forecast hours, which we denote as ${\bf x}$ and ${\bf y}$, respectively.
A conditional WGAN-GP, conditioned on the covariates ${\bf y_{cov}}$  in the HRDPS domain,  
aims to minimize the following losses for, respectively, the critic $\mathcal{L}_C$ and generator $\mathcal{L}_G$  \cite{AnnauEtAl2023Algorithmic, GulrajaniEtAl2017Improved, MirzaOsindero2014Conditional}:
\begin{equation}
    \mathcal{L}_C = \mathbb{E}_{G({\bf x}) \sim \mathbb{P}_g}[C(G({\bf x} \mid {\bf y_{cov}}))] - \mathbb{E}_{{\bf y} \sim \mathbb{P}_r}[C({\bf y})] + \lambda \mathbb{E}_{{\bf \hat{y}} \sim \mathbb{P}_{\hat{y}}}[(\| \nabla_{\hat{y}} C (\hat{y} )\|_2 -1)^2]
\end{equation}
and
\begin{equation}
    \mathcal{L}_G = - \gamma\mathbb{E}_{G({\bf x}) \sim \mathbb{P}_g}[C(G({\bf x} \mid {\bf y_{cov}}))] + \alpha \mathbb{E}_{{\bf y} \sim \mathbb{P}_r} l_c({\bf y}, G({\bf x} \mid {\bf y_{cov}})).
\end{equation}
In this formulation, \( C \) denotes a real-valued function known as the \emph{Critic}, which is defined on the high-resolution domain. The function \( G \) represents the \emph{Generator}, which maps low-resolution predictors to the high-resolution predictands. The symbol \(\mathbb{P}_g\) denotes the conditional distribution of the generated downscaled data given the high-resolution static covariates \({\bf y_{cov}}\), while \(\mathbb{P}_r\) represents the distribution of the high-resolution predictands.
The expression \(\mathbb{E}_{{\bf \hat{y}} \sim \mathbb{P}_{\hat{y}}}[(\| \nabla_{\hat{y}} C (\hat{y})\|_2 - 1)^2]\) represents the \emph{gradient penalty}, which is incorporated into the formulation to ensure valid Wasserstein distances by enforcing 1-Lipschitz continuity for the Critic. Here, \({\bf \hat{y}} \sim \mathbb{P}_{\hat{y}}\) denotes a linearly interpolated generated value defined by \({\bf \hat{y}} = \epsilon {\bf y} + (1 - \epsilon) G({\bf x} \mid {\bf y_{cov}})\), where \(\epsilon \sim U[0, 1]\). 
The scalars \(\lambda\), \(\gamma\), and \(\alpha\) are regularization hyperparameters: \(\lambda\) controls the gradient penalty term, \(\gamma\) the adversarial loss term, \(\mathbb{E}_{G({\bf x}) \sim \mathbb{P}_g}[C(G({\bf x} \mid {\bf y_{cov}}))]\) and \(\alpha\)  the content loss term, \(\mathbb{E}_{{\bf y} \sim \mathbb{P}_r} l_c({\bf y}, G({\bf x} \mid {\bf y_{cov}}))\).
All expectations in the equations above are computed across all dimensions of \({\bf x}\) and \({\bf y}\). Specifically, \({\bf x}\) is a tensor of size \(B \times C_{l} \times W_l \times H_l\), and \({\bf y}\) is a tensor of size \(B \times C_{h} \times W_h \times H_h\). Here, \(B\) represents the batch size, \(W\) and \(H\) denote the spatial width and height, respectively, while \(C\) indicates the number of input channels (variates). The subscripts \(l\) and \(h\) refer to the high- and low-resolution datasets, respectively. Additionally, \({\bf y_{cov}}\) is a tensor with dimensions \(C_{cov} \times W_l \times H_l\), where values are replicated across the batch size \(B\). This replication is due to the covariates being \emph{static} and independent of time. It is important to note that if \({\bf y_{cov}}\) is not provided, the WGAN-GP model reduces to the formulation presented in \cite{AnnauEtAl2023Algorithmic}.

Utilizing WGAN-GP for downscaling is preferred over traditional GANs due to the advantages offered by WGANs. These advantages include a more stable training process and the provision of meaningful loss metrics, which contribute to achieving better generative outputs. Additionally, WGAN-GP effectively mitigates common challenges encountered in GAN training, such as mode collapse and vanishing gradients. For a more comprehensive understanding of these concepts, please refer to the relevant literature \cite{AnnauEtAl2023Algorithmic, GulrajaniEtAl2017Improved, ArjovskyEtAl2017Wasserstein}.

\subsection{Frequency Separation}
The core idea of the frequency separation approach is to decompose the low- and high-frequency components and treat them differently during WGAN-GP training \cite{AnnauEtAl2023Algorithmic, FritscheEtAl2019Frequency}, with the aim of  enhancing the quality of downscaled wind components. This allows the adversarial training to focus specifically on generating the high-frequency components, which are crucial for capturing fine details.
In this approach, the high-frequency components are derived as \({\bf y}_{high} = {\bf y} - \mathscr{L}({\bf y})\) for the HRDPS data, and \(G({\bf x} \mid {\bf y_{cov}})_{high} = G({\bf x} \mid {\bf y_{cov}}) - \mathscr{L}(G({\bf x} \mid {\bf y_{cov}}))\) for the generated downscaled data, where \(\mathscr{L}\) represents a low-pass filter.
In this case, during WGAN-GP training, the Critic will focus on optimizing the loss based only in the high frequency components \({\bf y}_{high}\) and \(G({\bf x} \mid {\bf y_{cov}})_{high}\) instead of \({\bf y}\) and \(G({\bf x} \mid {\bf y_{cov}})\). 
On the other side the Generator delegates the generation of high frequency components to the adversarial loss component 
\( - \mathbb{E}_{G({\bf x})_{high} \sim \mathbb{P}_g}[C(G({\bf x} \mid {\bf y_{cov}})_{high})] \) and the low frequency components to the content loss \(\mathbb{E}_{{\bf y} \sim \mathbb{P}_r} l_c( \mathscr{L}({\bf y}), \mathscr{L}( G({\bf x} \mid {\bf y_{cov}}))) \).\footnote{Annau et.al.,  proposed partial frequency separation \cite{AnnauEtAl2023Algorithmic},  this approach is similar to  frequency separation, but the adversarial loss is used in its original formulation, i.e., \( - \mathbb{E}_{G({\bf x}) \sim \mathbb{P}_g}[C(G({\bf x} \mid {\bf y_{cov}}))] \).}

\subsection{Architectures}
\subsubsection{Critic Network}
\begin{figure}[ht]
    \centering
    \includegraphics[width=1\textwidth]{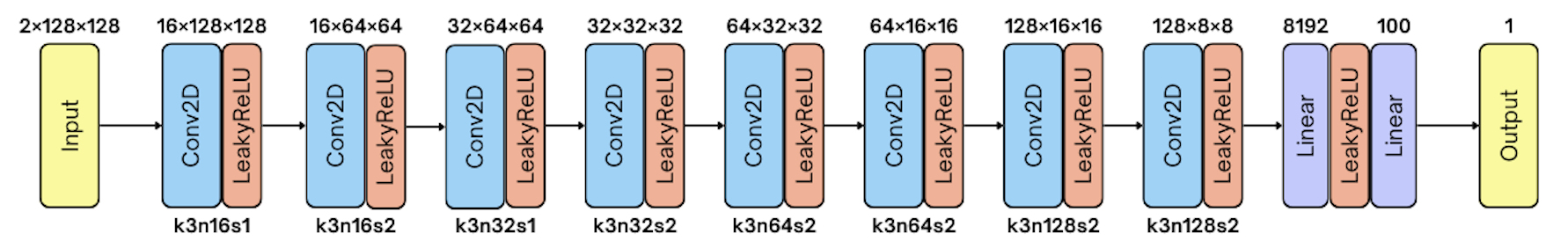}
    \caption{Schema of the Critic Network. The input consists of either a block of high-resolution NWP predictands or downscaled data generated by the Generator. The output is a real-valued score representing the Critic's confidence that the input originates from actual NWP data—a higher score indicates a more physically realistic appearance. The labels $k$, $n$ and $s$ below the blocks denote the kernel size, the number of filters, and the stride used in each layer, respectively. The numbers above the blocks represent the tensor dimensions as a function of the number of feature maps, width, and height.}
    \label{fig:critic}
\end{figure}

To implement the proposed conditional WGAN-GP, we employed the same Critic network described by \cite{AnnauEtAl2023Algorithmic}, as illustrated in Figure~\ref{fig:critic}. The figure provides a schematic of the Critic network, with the numbers above each block indicating the dimensions of the data representations at various stages. The labels below the blocks denote the kernel size $k$, the number of filters $n$, and the stride $s$ used in each layer. The Critic utilizes a VGG-style architecture that progressively reduces the spatial dimensions of the input \cite{simonyan2014very}, which may include HRDPS wind components or the downscaled version of GDPS wind components provided by the Generator network, through a sequence of convolutional layers. The network processes the input data using 3×3 convolutions with LeakyReLU activations\footnote{LeakyReLU are defined as:  \(\sigma(x) = \max(x, \alpha x)\)
where \( x \) is the input to the activation function, and \( \alpha \) is a small positive parameter that determines the slope for negative inputs.}, where each layer increases the number of feature maps while downsampling the spatial dimensions using a stride of 2. This approach allows the network to expand the number of channels up to eight times the input dimension. The extracted features are then passed through a final processing block, a regression-style scoring mechanism that reduces the feature vector to a single scalar output. This scalar score, used as an approximation of the Wasserstein distance, indicates the Critic's assessment of the physical realism of the generated data. 

\subsubsection{Generator Network}
\begin{figure}[ht]
    \centering
    \includegraphics[width=1\textwidth]{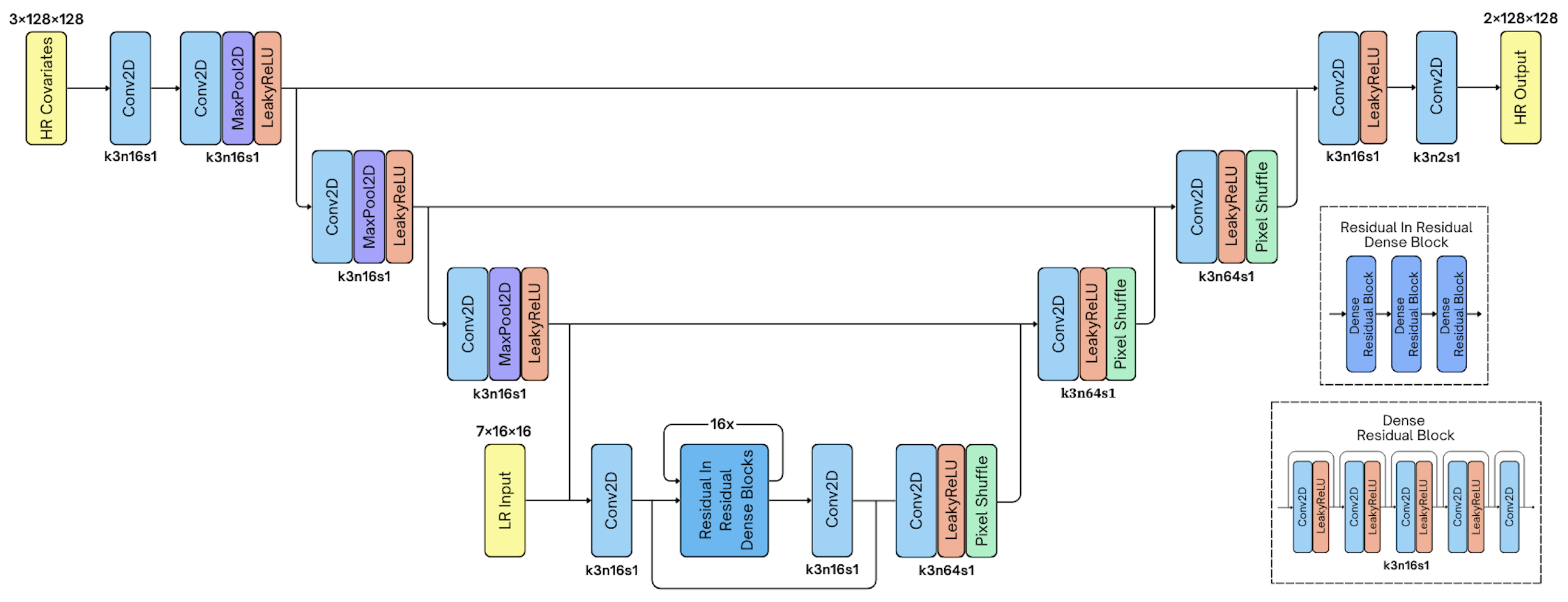}
    \caption{Schematic of the Generator Network. The inputs consist of static high-resolution covariates (top-left) and low-resolution predictors (buttom-left). The static covariates go through three successive coarsening passes and are then concatenated with the low-resolution predictors. The resulting combination then goes sixteen times through the core component called "residual in residual dense block". Three refinement passes with skip connections followed by a last group of convolution yield the final high-resolution output.}
    \label{fig:unet}
\end{figure}

The generator network implemented in the proposed model draws upon concepts 
from \cite{AnnauEtAl2023Algorithmic, LedigEtAl2017Photorealistic} and is 
structured according to a U-Net architecture.
The architecture is depicted in Figure \ref{fig:unet}.
We employ this design because it facilitates the integration of static HRDPS covariates 
and GDPS variates as inputs. 
The Generator embeds the HRDPS covariates through a sequence of downsampling layers 
reducing the spatial dimensions by a factor of eight. 
The resulting output tensor is then concatenated with the feature map tensor derived from the 
GDPS variates.
This combined tensor passes through a series of residual-in-residual dense blocks, 
for a deep feature extraction. 
Next, the network applies a series of upsampling layers that leverage 
pixel shuffle operations~\cite{shi2016realtimesingleimagevideo} to progressively 
restore spatial resolution. 
The pixel shuffle operation restructures the input tensor’s feature maps into a higher-resolution grid, 
effectively increasing spatial resolution. 
For an upscale factor of 2, the operation requires four times the number of 
input feature maps to redistribute them into the upsampled spatial dimensions 
while maintaining the number of output channels.
Figure \ref{fig:unet} shows an $n= 64$ ($n64$ below each upsampling block), 
representing 64 input feature maps required to produce an upsampled output tensor 
with 16 feature maps.
Throughout the upsampling process, skip connections from the downsampling layers are 
leveraged to facilitate the merging of local and global features while 
preserving high-frequency details.
The final output layer incorporates a convolutional operation to generate 
the GDPS downscaled version of  wind components. 
This U-Net architecture serves as the mechanism for 
implementing the conditional Wasserstein GAN, 
effectively utilizing static HRDPS covariates to guide the generative process of 
downscaled GDPS wind components.

\section{Experiments}
\subsection{Dataset Split}\label{sec:datasetsplit}
This section describes the experimental setup, using a dataset of matched GDPS and HRDPS forecast hours. 
For each prediction system, 24 hours of forecasts are used for each day consisting of forecast lead-times 6-17 hours for each of the 00 and 12 UTC runs. In this way, we avoid both the model spin-up time, and later forecast times where the two systems might have diverged because of error growth.
For training, we selected all GDPS and HRDPS files with predictions starting between July 2022 to June 2023. The dataset was randomly split into 7 150 files for training and  1 178 for validation.
We use an independent dataset, separate from the training and validation intervals, as the test set. It comprises forecast lead times from 6 to 17 hours for July 2023 (682 files), August 2023 (682 files), January 2024 (618 files), and February 2024 (696 files).

\subsection{Metrics}  
To evaluate the performance of the proposed conditional WGAN-GP, 
we used the Root Mean Squared Error (RMSE) between the HRDPS reference and the downscaled GDPS wind components. 
Additionally, we employed the
Radially Averaged Log-Spectrum Distance (LSD)~\cite{HarrisEtAl2022Generative}:

\begin{equation} 
    LSD(\overline{P}_{ref}, \overline{P}_{pred}) = \sqrt{\frac{1}{N_r} \sum_{r} \left( 10 \log_{10} \frac{\overline{P}_{ref}(r)}{\overline{P}_{pred}(r)} \right)^2},
\end{equation}
where \( P(r) \) is the Radially Averaged Power Spectrum Density (RAPSD) \cite{HarrisEtAl2022Generative, ruzanski2011scale} 
Observe that LSD quantifies the difference between predicted and true power spectra on a logarithmic scale. Subscripts \( _{pred} \) and \( _{ref} \) are  the \( P(r) \) of predicted and reference wind fields, respectively.

\subsection{Training Procedures}

The regridded GDPS (see Section 
\ref{sec:preprocessing}) and HRDPS datasets cover the entire Canadian region (except the Arctic Archipelago) with a grid of \((2540, 1280)\) spatial values.
In this sense, each GDPS and HRDPS file is represented by tensors of sizes \((7, 2540, 1280)\) and \((2, 2540, 1280)\), 
where 7 and 2 represent the variable count in GDPS and HRDPS, respectively, as detailed in Table \ref{table:vars_description}.
To train the WGAN-GP, we based our implementation in the open-source implementation from \cite{Annau2022DoWnGANa}. We used  the following hyperparameters:
\emph{critic iterations}: 5 (indicating one generator update per five critic updates), 
\emph{batch size}: 32,  \emph{learning rate}: 0.00025, \emph{regularization hyperparameters}:   \(\lambda = 10\) (gradient penalty), \(\gamma = 0.01\) (adversarial loss), and \(\alpha = 5\) (content loss).
We used spatial random cropping for training with a \((128 \times 128)\) crop size and a downsampling factor of 8 for the regridded GDPS, 
resulting in GDPS tensors of \((7, 16, 16)\) for each corresponding HRDPS tensor of \((2, 128, 128)\).
We generated 832 crops per GDPS/HRDPS pair, amounting to 5\,948\,800 random crops for the 7\,150 pairs in the training dataset. Here, one "training step" corresponds to using all 832 random crops from a single GDPS-HRDPS forecast hour. This random crop setup allows six generator updates using a batch size of 32, fitting in the 40 GB of RAM of a Nvidia's A100 GPU chip we used for training.

The UNET generator network is configured with inputs of \((32, 7, 16, 16)\) (batch size, variables, spatial dimensions) and receives an HRDPS static covariate tensor of \((3, 128, 128)\). The generator produces  tensors of \((32, 2, 128, 128)\), representing the downscaled u10 and v10 wind components in the HRDPS domain. The critic network processes input tensors of \((32, 2, 128, 128)\), derived from either HRDPS data or the downscaled GDPS output.
 The proposed conditional WGAN-GP was trained for 65,250 training steps, with validation involving 32 random crops per GDPS-HRDPS pair in the validation set. 

\subsection{Effect of using HRDPS Static Covariates}
In this experiment, we investigate the impact of incorporating HRDPS static covariates on downscaling accuracy. To this end, we compare the DownGAN model \cite{AnnauEtAl2023Algorithmic} using the open-source implementation from \cite{Annau2022DoWnGANa}, against the proposed conditional WGAN-GP that utilizes HRDPS static covariates. Both models are trained with identical hyperparameters and configurations to ensure a fair comparison. The primary difference lies in their generator architectures: DownGAN employs a generator based on the SRGAN network introduced by \cite{LedigEtAl2017Photorealistic}, while the WGAN-GP model adapts the DownGAN generator to a UNET structure, allowing it to effectively incorporate the additional HRDPS covariates.

\begin{figure}[ht]
    \centering
    \includegraphics[width=0.5\textwidth]{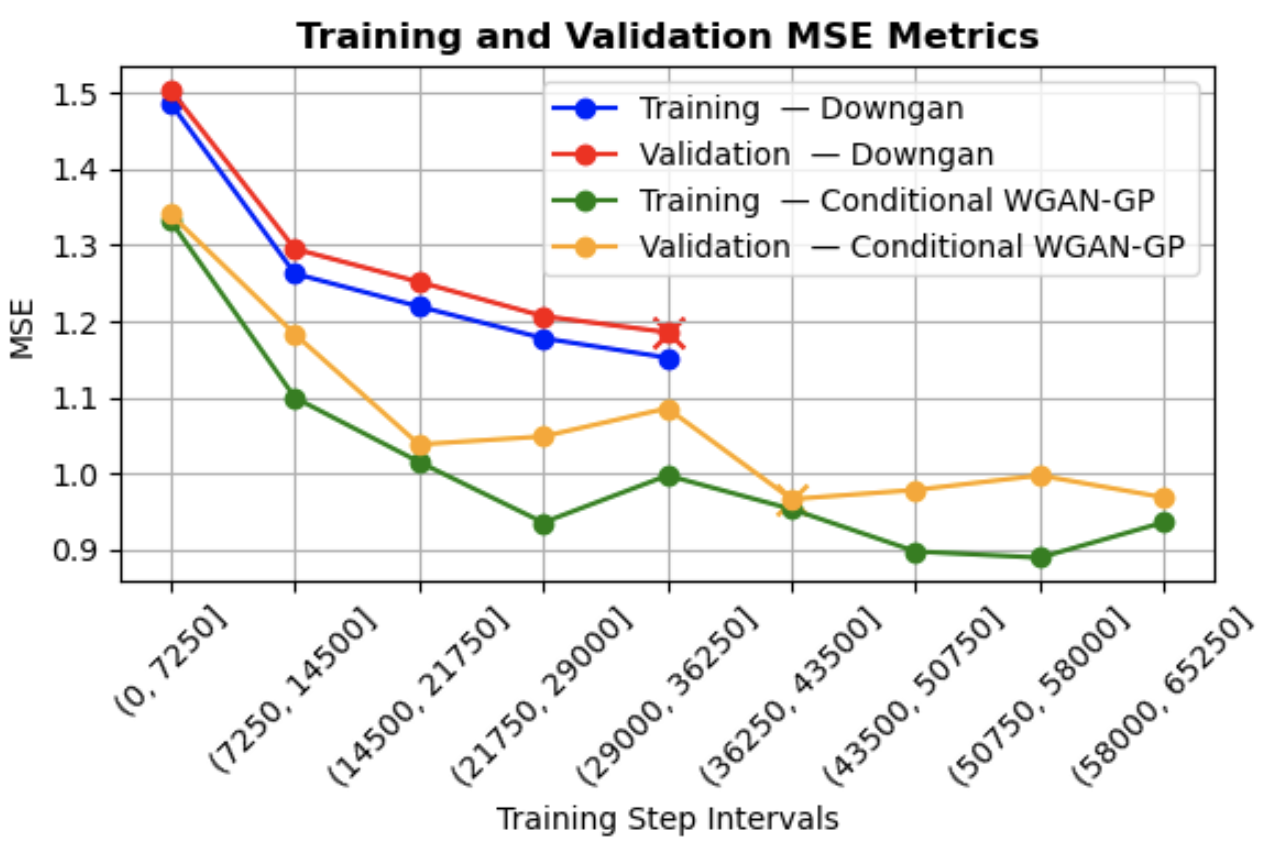}
    \caption{Training and validation curves: DownGAN vs  conditional WGAN-GP. Conditional WGAN-GP outperforms DownGAN in terms of the validation MSE curve and achieves faster convergence during training.}
    \label{fig:effect_covariates}
\end{figure}
Figure \ref{fig:effect_covariates} presents the MSE curve trajectories for the DownGAN and the  Conditional WGAN-GP models
across training and validation steps. 
The MSE values are averaged over intervals of 7\,250 training steps, which correspond to the number of files in the training set.
In the plot, the DownGAN model's training and validation MSE average values are represented by blue and red markers, respectively, while the Conditional WGAN-GP model's MSE average values are shown in green (training) and orange (validation). 
The Conditional WGAN-GP model, which incorporates HRDPS static covariates, demonstrates consistently lower MSE values compared to the DownGAN model.
The minimum validation MSE average value for each model is marked with an "x". The Conditional WGAN-GP model attains a lower validation error across all intervals, suggesting its enhanced capability for accurately downscaling surface winds. 

\subsection{Effect of Frequency Separation}
\label{sec:EFS}
\begin{figure}[ht]
    \centering
\includegraphics[width=0.85\textwidth]{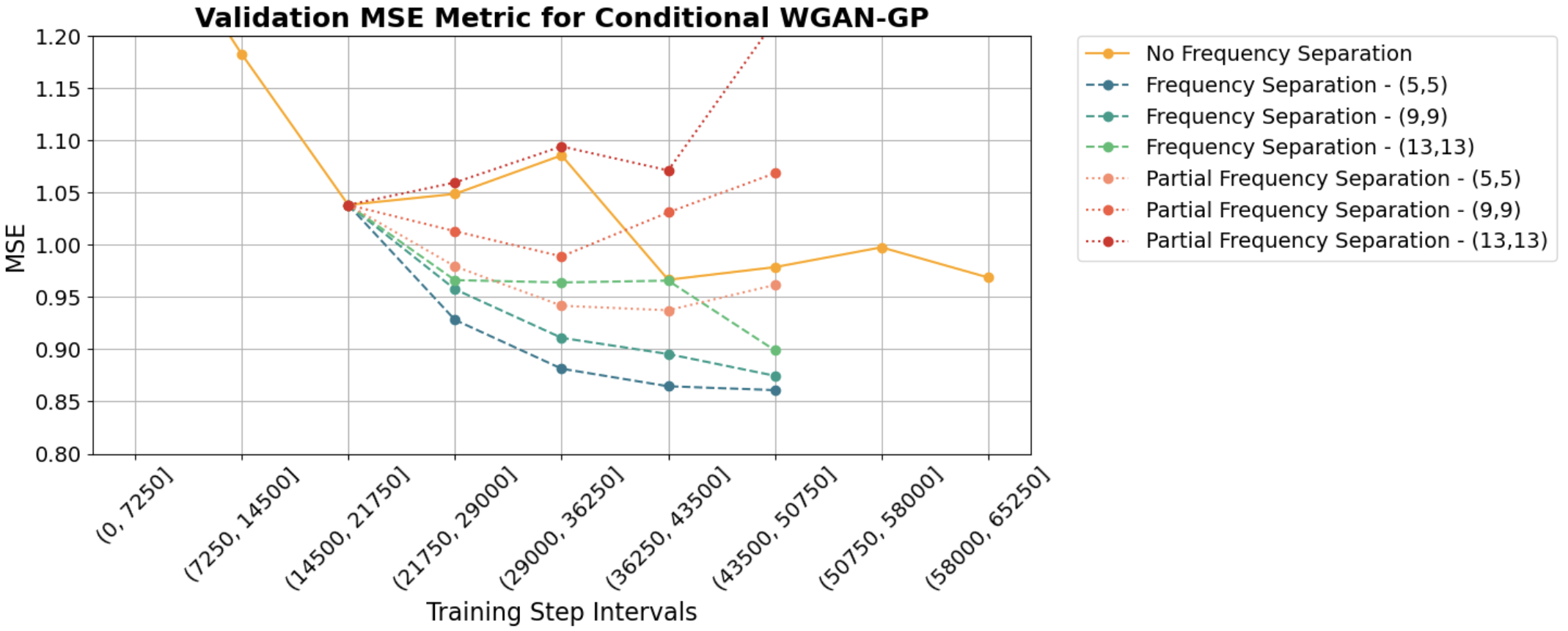}
    \caption{Effect of Frequency separation in the validation curve. Frequency separation with filter sizes of (5,5), (9,9) or (13,13) has a positive effect on the MSE validation score (0.86, 0.87 and 0.9, respectively), but partial frequency separation does not improve the scores or only slightly (MSE validation score of 0.93) for filter size of (5,5).}
    \label{fig:effect_fs}
\end{figure}

This section shows the experimental results to investigate the impact of Frequency Separation (FS) and Partial Frequency Separation (PFS) on the performance of the conditional WGAN-GP model. We tested multiple configurations  for both FS and PFS, varying the low-pass average filter kernel sizes. Specifically, we used kernel sizes of \((5,5),  (9,9)\) and \((13,13)\) for each WGAN-GP model with FS and PFS. Additionally, we included a conditional WGAN-GP model without FS or PFS as a baseline (i.e., same MSE validation curve from Figure \ref{fig:effect_covariates}).
For the experiments, we applied transfer learning  starting from training step 21\,750, and continuing training up to step 50\,750. Figure \ref{fig:effect_fs} presents a comparison of the models’ performance based on the MSE validation metric.

From the experiments, we observed that FS is in general better than PFS, and than small filter kernel sizes are better in term of the validation MSE metric.
For this experiment, WGAN-GP with FS achieved the best configuration using a filter size of \((5,5)\), obtaining an average MSE validation score of 0.86 in the interval \([43 500, 50 750]\). This result outperformed both the best PFS configuration (filter size of \((5,5)\) and MSE validation score of 0.93)  and the case without frequency separation (MSE validation score of 0.97).

\begin{figure}[ht]
    \centering
\includegraphics[width=0.8\textwidth]{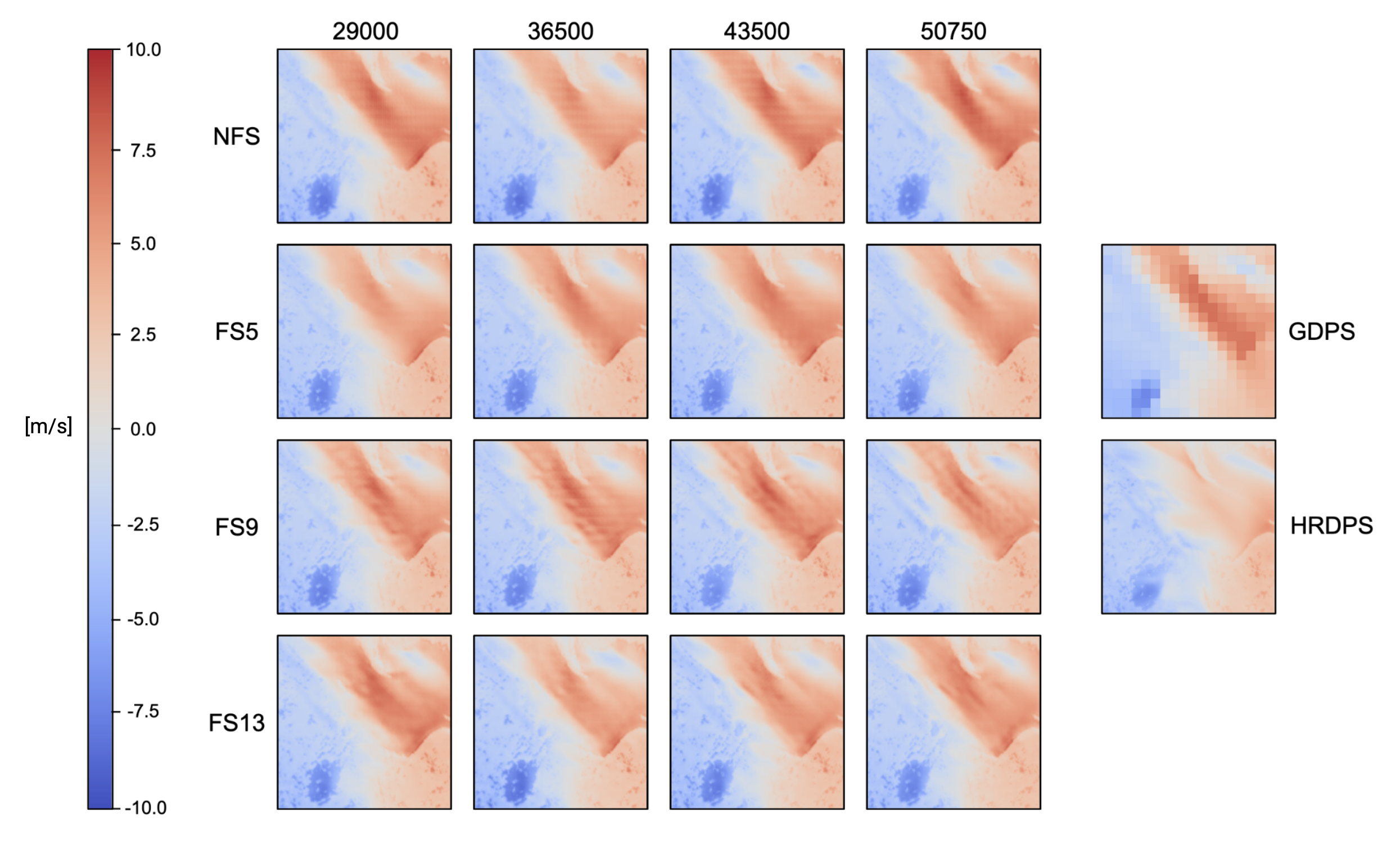}
    \caption{Effect of frequency separation on downscaling the v10 wind component at 15:00 UTC on July 1, 2023, at training steps 29\,000, 36\,500, 43\,500, and 50\,750. Rows show different configurations; columns show training steps. Corresponding wind fields from GDPS and HRDPS are shown for reference. All the downscaled fields capture well the main features of the HRDPS forecast, but frequency separation reduces checkboard artifacts and a larger filter size leads to a sharper forecast.}
    \label{fig:effect_fs_pic}
\end{figure}
Figure \ref{fig:effect_fs_pic} shows the effect of FS in dealing with high frequencies as a function of training steps 29\,000, 36\,250, 43\,500 and 50\,750. In this case, checkboard artifacts are less evident in WGAN-GP with FS than in the case with no frequency separation (NFS), which can be explained by the greater focus on high-frequency components when training with FS.

\subsection{Metrics on the Test Set}
\begin{figure} \centering \includegraphics[width=1\textwidth]{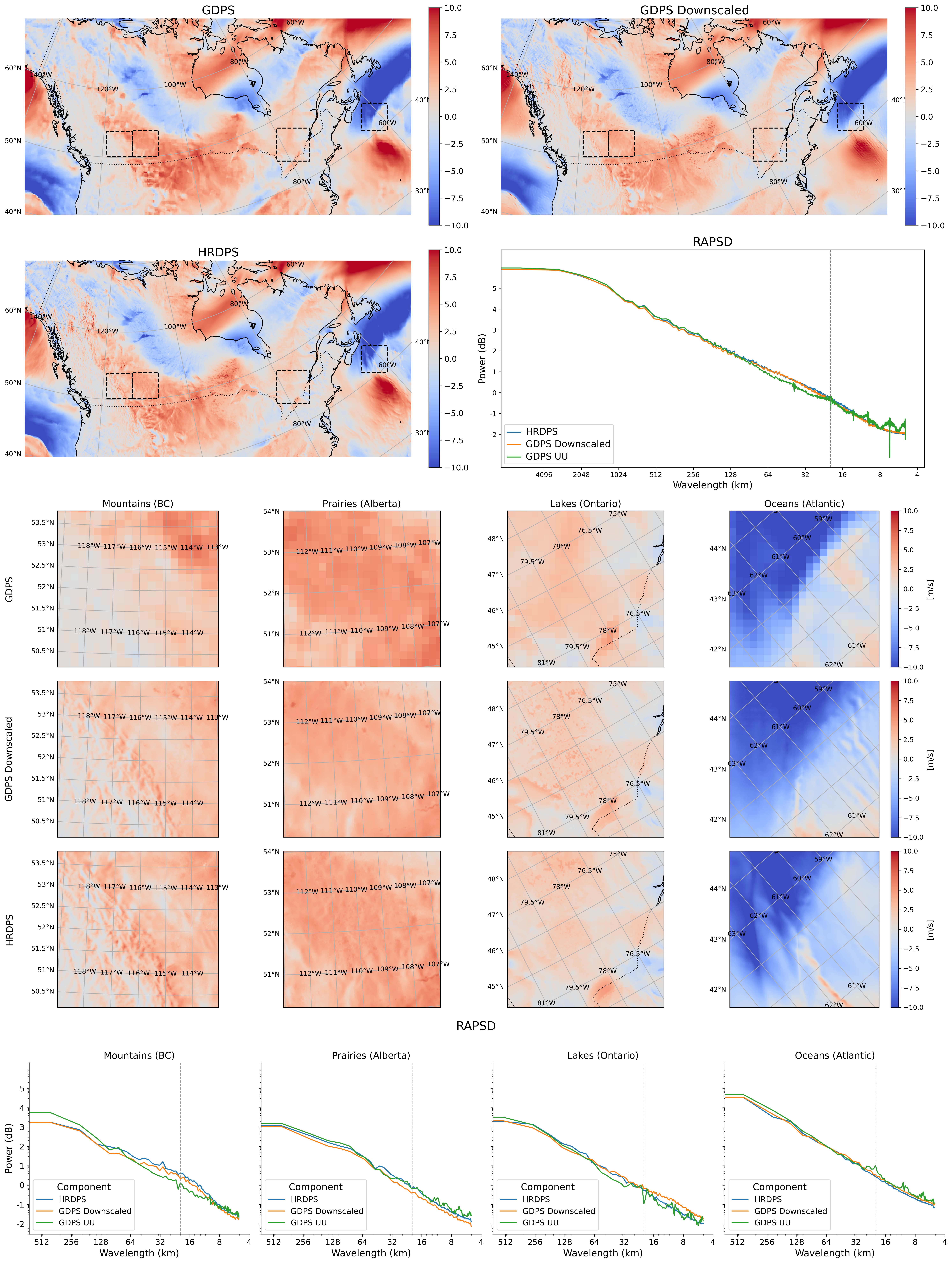} \caption{GDPS, HRDPS, and GDPS downscaled forecasts for the test set (u10 wind component) corresponding to the 12:00 UTC run at 06:00 on February 22, 2024. The downscaling was performed over the entire Canadian domain using the WGAN-GP model with frequency separation and a filter size of (13,13), utilizing trained model checkpoint 50\,750. 
The RAPSD graph show the RAPSD plots for HRDPS  the downscaled GDPS, and the GDPS, here
the gray dashed vertical line describe the cut frequency for the GPDS \(20 km^{-1}\).
Bounding boxes in black over GDPS are visualized in bottom part, for four distinct regions: Mountains, Prairies, Lakes, and Oceans. This figure highlights how the AI downscaling model captures fine-scale features in areas with strong topographic influence while revealing its limitations in regions with no topographic information, such as the oceans.
} 
\label{fig:DownscaleWholeRegion} 
\end{figure}

This section presents the metrics computed on the independent test set described in Section \ref{sec:dataset}.
We compute the metrics by performing downscaling over the entire Canadian domain using the trained generator. Although the models were trained with random 16 × 16 crops to produce images of 128 x 128, we use the trained generator to perform downscaling on the entire Canadian region. Specifically, we send the full GDPS file, which covers the entire Canadian domain, directly to the generator without the need to manually stitch together the 128 x 128 blocks\footnote{This is possible because a convolutional layer's parameter count depends only on the filter size and the number of input and output feature maps, remaining unaffected by input width and height. This enables convolutional networks to handle varying spatial dimensions without changing the number of trainable parameters.}. To ensure compatibility with the network's design for an exact 8x resolution increase, we adjust the original dimensions. The original GDPS input dimensions (162, 318) are modified to (161, 317), producing a downscaled GDPS output with dimensions of (1288, 2536). Similarly, we adjust the original HRDPS reference dimensions from (1290, 2540) to (1288, 2536) to enable direct comparison with the downscaled GDPS version.
Figure \ref{fig:DownscaleWholeRegion}  presents an example of downscaling the 
u10 wind component from the GDPS hourly forecast from the test set, corresponding to the 12:00 UTC run at 06:00 on February 22, 2024, over the entire Canadian domain,  using the WGAN-GP model with a filter size of (13,13) and checkpoint 50\,750. In this case, the RMSE between the downscaled GDPS and HRDPS is 0.842, while the log spectral distance (LSD) is 0.655. Subjective assessment of the downscaled GDPS confirms that overall the output matches closely the HRDPS and enhances the details in mountainous areas and sharpen the boundary between weak and strong winds over lakes. However, some details are still smoothed or mismatched over the two oceans (Atlantic and Pacific). The use of static covariates thus enable to effectively downscale topographically-driven features, but in the absence of strong topographic features such as over the oceans and in the prairies, the AI downscaling model only partially reconstructs high-frequency details. Over the mountains, the power spectra match very well, but over the prairies the power spectra of the downscaled GDPS loses some energy below 50 km resolution. Over the oceans, the detailed locations of the wind features are wrong even if the features themselves look fairly realistic. The power spectra thus match very well even if the wind patterns are different. For this particular winter case, the wind over lakes is stronger for the downscaled GDPS than for the target HRDPS, possibly because a failure of the downscaling model to account for a change of surface roughness length when lakes ice. It is reflected with a higher energy power spectra at higher frequencies for the downscaled GDPS compared to the target HRDPS.

\begin{figure}[!htbp]
    \centering
    \includegraphics[width=1\textwidth]{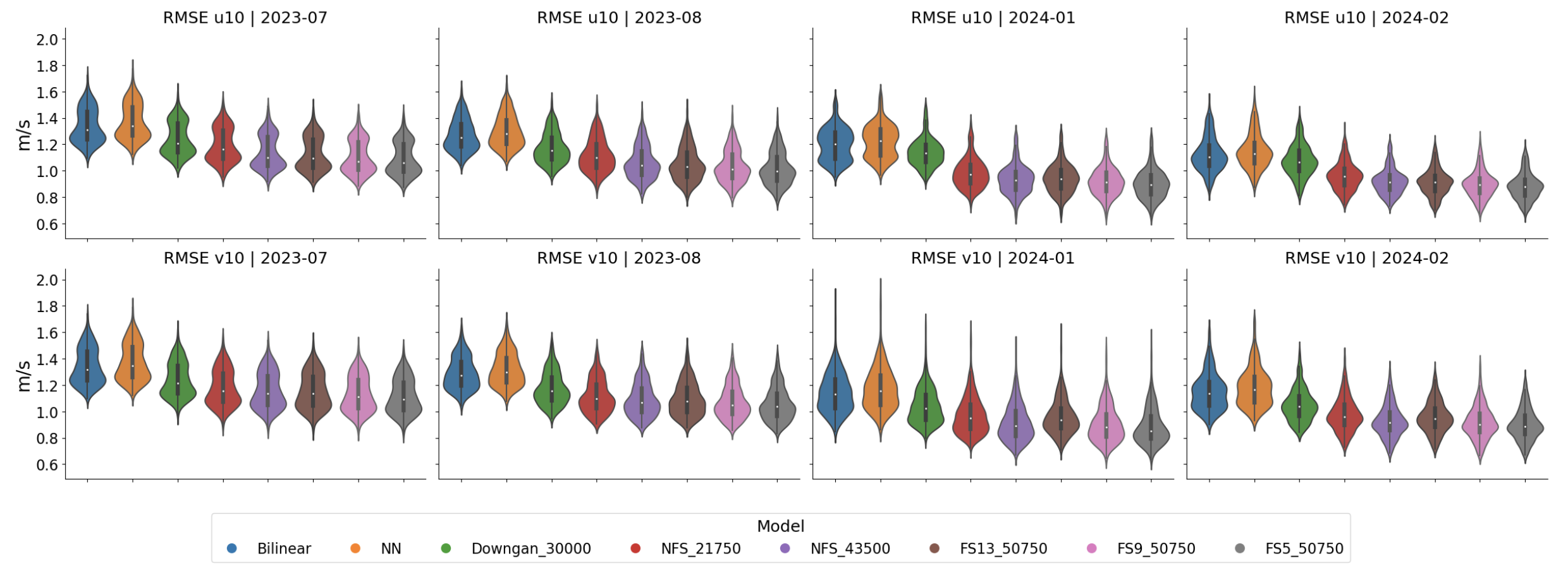}
    \caption{Violin plots of the RMSE of the u10 and v10 on the test sets for four different months: July 2023, August 2023, January 2024 and February 2024. Simple interpolation such as bilinear and nearest neighbour performs worst for all months and wind components. The DownGAN model performs slightly better, but the Conditional WGAN-GP is best. Training for 43\,500 steps is slightly better than for 21750 steps and adding frequency separation further enhances the scores. Overall, RMSE is lower for Winter months (January-February) than for Summer months (July-August) and similar for u10 and v10 wind components.}
    \label{fig:metrics_test_Set_RMSE}
\end{figure}

\begin{figure}[!htbp]
    \centering
    \includegraphics[width=1\textwidth]{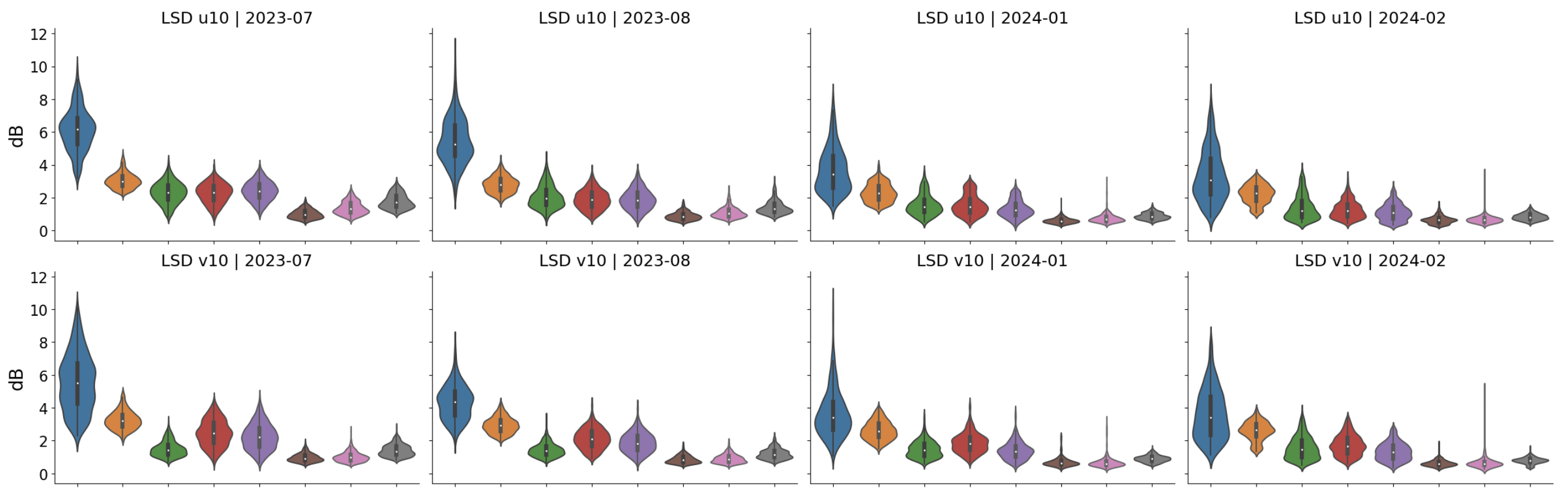}
    \caption{Violon plots of LSD on the test sets for the same months as Fig.\,\ref{fig:metrics_test_Set_RMSE}. Bilinear interpolation has clearly the worst power spectrum for all months and wind components followed by the nearest neighbor interpolation. DownGAN does similar than the Conditional WGAN-GP without frequency separation, except for v10 in Summer months where it does better. However, the power spectra that are the most similar to the HRDPS power spectra are the Conditional WGAN-GP with frequency separation. Frequency separation with large filter size of (13,13) is better than with smaller filter size of (5,5), particularly for July and August.}
    \label{fig:metrics_test_Set_LSD}
\end{figure}

Figures \ref{fig:metrics_test_Set_RMSE} and \ref{fig:metrics_test_Set_LSD} shows the distribution of RMSE and LSD values on the test set, comparing downscaled GDPS forecasts against HRDPS references. The evaluation is conducted separately for the u10  and v10 components and further stratified by month.  
Bilinear interpolation and nearest-neighbor (NN) interpolation serve as baseline methods. In addition, we assess multiple models, including the DownGAN model at checkpoint 29\,000 and the Conditional WGAN-GP model under different configurations: this model is evaluated at checkpoints 21\,750 and 43\,500 in the absence of frequency separation (denoted as NFS-21\,750 and NFS-43\,500, respectively) and with frequency separation using filter sizes of \( (13,13) \), \( (9,9) \), and \( (5,5) \) at checkpoint 50\,750 (denoted as FS13-50\,750, FS9-50\,750, and FS5-50\,750, respectively).  
The selection of model checkpoints is guided by the lowest validation MSE observed in the experiments detailed in Section \ref{sec:EFS}.

Analyzing the RMSE distributions in Figure \ref{fig:metrics_test_Set_RMSE} for both wind components, u10 and v10, we observed that the interpolation methods (Bilinear and NN) produced the highest RMSE values, indicating their inferior performance. The DownGAN model also yielded higher RMSE values compared to the Conditional WGAN-GP variations. Among the Conditional WGAN-GP models, those incorporating frequency separation achieved the lowest RMSE, demonstrating superior performance in this metric.

The case of LSD distributions shown in Figure \ref{fig:metrics_test_Set_LSD} highlights the role of the filter size in the Conditional WGAN-GP with Frequency Separation. In this case, a filter size of \(13, 13\) or \(9, 9\) achieved the lowest LSD metric, indicating its superior performance according to this metric.

Table \ref{tab:metrics_summary} presents a summary of the metrics in the test set, aggregating all the months,  in terms of their median and median absolute deviation (MAD) statistics.  Marked in bold is the best model for RMSE and for LSD, for both u10 and v10. Overall, the Conditional WGAN-GP with frequency separation with a filter size of (5,5) has the lowest RMSE for both u10 and v10, but increasing the filter size to (9,9) or (13,13) leads to, respectively, the lowest LSD for v10 and u10. Note that the difference in median RMSE between these three models is only 0.04 m/s, which is smaller than the MAD of 0.09 m/s. However, the difference of LSD for u10 is 0.38 dB, which is more than the MAD of 0.18 dB for FS13. Similarly the difference of LSD for v10 is 0.16 dB which is about the same as the MAD of F13, but the difference between FS9 and FS13 is only 0.004 dB. The F13 model offers the best balance between RMSE and LSD performance.

\footnotesize
\begin{longtable}{lcccccccc}
\caption{Summary of results. } \label{tab:metrics_summary} \\
\toprule
\textbf{Method} & \multicolumn{2}{c}{\textbf{RMSE (u10)}} & \multicolumn{2}{c}{\textbf{RMSE (v10)}} & \multicolumn{2}{c}{\textbf{LSD (u10)}} & \multicolumn{2}{c}{\textbf{LSD (v10)}} \\
\cmidrule(lr){2-3} \cmidrule(lr){4-5} \cmidrule(lr){6-7} \cmidrule(lr){8-9}
 & \textbf{Median} & \textbf{MAD} & \textbf{Median} & \textbf{MAD} & \textbf{Median} & \textbf{MAD} & \textbf{Median} & \textbf{MAD} \\
\midrule
\endhead
\bottomrule
\endfoot
Bilinear & 1.224 & 0.103 & 1.220 & 0.107 & 4.703 & 1.418 & 4.124 & 1.083 \\
NN & 1.252 & 0.103 & 1.252 & 0.110 & 2.625 & 0.417 & 2.843 & 0.398 \\
DownGAN-30000 & 1.142 & 0.080 & 1.116 & 0.100 & 1.743 & 0.566 & 1.408 & 0.334 \\
FS13-50750 & 0.993 & 0.086 & 1.023 & 0.100 & {\bf 0.744} & 0.181 & 0.735 & 0.164 \\
FS9-50750 & 0.974 & 0.087 & 0.995 & 0.109 & 0.877 & 0.253 & {\bf 0.731} & 0.188 \\
FS5-50750 & {\bf 0.959} & 0.089 & {\bf 0.981} & 0.111 & 1.124 & 0.315 & 0.998 & 0.220 \\
NFS-21750 & 1.048 & 0.097 & 1.051 & 0.106 & 1.689 & 0.539 & 1.985 & 0.498 \\
NFS-43500 & 0.997 & 0.090 & 1.015 & 0.114 & 1.624 & 0.570 & 1.609 & 0.488 \\
\end{longtable}
\normalsize 

\section{Conclusions}
In this study, we presented a downscaling methodology based on a conditional Wasserstein GAN (WGAN-GP), leveraging low-resolution data from the GDPS and high-resolution data from the HRDPS, conditioned on high-resolution static covariates. We also conducted experiments incorporating the Frequency Separation approach to enhance the performance of the model. To implement the conditional WGAN-GP, we employed a UNET model as the generator. This UNET was designed to process two types of inputs: high-resolution static covariates and low-resolution GDPS data. Due to memory constraints, the training protocol relied on extracting random crops from single pairs of GDPS and HRDPS data. We carried out experiments to analyze the impact of including high-resolution static covariates in the model. Our findings indicate that these covariates significantly improved model convergence. Furthermore, experimental results demonstrated that the use of Frequency Separation contributed to further performance gains. We evaluated the generalization capability of the proposed model using a held-out test set. The results revealed that the best model configuration incorporated Frequency Separation, achieving favorable performance in terms of RMSE and Log Spectral Distance (LSD) metrics.

In preparation for an operational environment, we trained the AI model on small \(128 \times 128 \) image patches using a downsampling factor of 8 and made inferences over the entire Canadian domain \(1288 \times 2536 \). As future work, we aim to develop the model as a product capable of extending high-resolution wind forecast services from the current 48-hour horizon to the 10-day forecast horizon of the Canadian global forecast model. Additionally, we plan to investigate the application of recent state-of-the-art weather foundation models \cite{SchmudeEtAl2024Prithvi} for wind downscaling.


\bibliographystyle{unsrt}
\bibliography{references2}
\end{document}